\documentclass{article}

\usepackage[preprint]{neurips_2021}

\usepackage[utf8]{inputenc} 
\usepackage[T1]{fontenc}    
\usepackage{hyperref}       
\usepackage{url}            
\usepackage{booktabs}       
\usepackage{amsfonts}       
\usepackage{nicefrac}       
\usepackage{microtype}      
\usepackage{xcolor}         
\usepackage{amsmath}
\usepackage{graphicx}
\usepackage{algorithmic}
\usepackage{algorithm}
\usepackage{url}
\usepackage{multirow}
\usepackage{siunitx}
\usepackage{caption}
\usepackage{subcaption}

\title{Generative Adversarial Imitation Learning for Empathy-based AI}

\author{
  Pratyush Muthukumar\\
  Information and Computer Science\\
  UC Irvine\\
  Irvine, CA 92697 \\
  \texttt{muthukup@uci.edu} \\
   \And
   Karishma Muthukumar \\
   Social Sciences \\
   UC Irvine \\
  \texttt{muthukuk@uci.edu} \\
   \AND
   Deepan Muthirayan \\
   Electrical Engineering and Computer Science \\
   UC Irvine \\
   \texttt{dmuthira@uci.edu} \\
   \And
   Pramod Khargonekar \\
   Electrical Engineering and Computer Science \\
   UC Irvine \\
   \texttt{pramod.khargonekar@uci.edu} \\
}

\begin{document}

\maketitle

\begin{abstract}
  Generative adversarial imitation learning (GAIL) is a model-free algorithm that has been shown to provide strong results in imitating complex behaviors in high-dimensional environments. In this paper, we utilize the GAIL model for text generation to develop empathy-based context-aware conversational AI. Our model uses an expert trajectory of empathetic prompt-response dialogues which can accurately exhibit the correct empathetic emotion when generating a response. The Generator of the GAIL model uses the  GPT-2 sequential pre-trained language model trained on 117 million parameters from 40 GB of internet data. We propose a novel application of an approach used in transfer learning to fine tune the GPT-2 model in order to generate concise, user-specific empathetic responses validated against the Discriminator. Our novel GAIL model utilizes a sentiment analysis history-based reinforcement learning approach to empathetically respond to human interactions in a personalized manner. We find that our model’s response scores on various human-generated prompts collected from the Facebook Empathetic Dialogues dataset outperform baseline counterparts. Moreover, our model improves upon various history-based conversational AI models developed recently, as our model’s performance over a sustained conversation of 3 or more interactions outperform similar conversational AI models.
\end{abstract}

\section{Introduction} \label{int}

Text generation models designed for intelligent conversation systems are increasingly popular in research. Many modern text generation models take advantage of deep learning paradigms for natural dialogue generation. State-of-the-art results in this field utilize the numerous processing layers of deep learning architectures typified by Recurrent Neural Networks, Variational Autoencoders, and Transformer models \citep{sutskever2011generating, zhang2019improve, li2020transformer}. These neural network approaches are capable of context-aware, history-based text generation, however, it is less often the case we see research focused on developing empathetic dialogue systems. 

The challenge of developing empathetic dialogue systems is inherently a conceptual one: in literature and application, there is no clear definition of empathy. Cognitive neuroscientists are actively seeking to define empathy empirically, however, even the most objective definitions fail to cater to each individual's perception of empathy \citep{gerdes2010conceptualising}. It is even more so challenging to develop artificial empathy capable of detecting and responding to human emotions \citep{asada2015towards}.

In this paper, we propose an architecture capable of empathetic, context-aware, natural dialogue generation. To tackle the challenges of such a problem, we employ the use of deep reinforcement learning. Deep reinforcement learning has recently been used for traditional text generation, as the nature of reinforcement learning allows for success in complex, high dimensional environments. The challenge of picking an appropriate response out of an entire set of possible utterances in a language is certainly the caliber of problem that reinforcement learning is able to solve. \citet{li2016deep} proposed one of the first applications of deep Q-learning for dialogue generation by simulating dialogue between two virtual agents using policy gradient methods. Their work showed improvements over traditional MLE-based Seq2Seq models, citing the main advantages of deep Q-learning to be its ability to keep the conversation moving through non-generic responses and its ability to recognize repetitive utterances. 

While traditional reinforcement learning has shown success for natural dialogue generation, a key shortcoming is evident when we introduce empathy-based intelligent conversation agents. For an empathetic dialogue agent to select an optimal policy in a state-space, it must maximize a reward function which defines empathetic conversation. However, the same problem arises: it is virtually impossible to define a statistical formula for empathy. 

We propose the application of inverse reinforcement learning or imitation learning to solve this problem in a different way. Instead of postulating an empirical formula for empathy, we feed the model a trajectory of empathetic expert actions that an empathetic dialogue agent can imitate. Our rationale behind this approach is that it is significantly easier to recognize examples of empathy in conversation, actions, or people rather than developing an overarching formula for empathy. 

In our implementation, we utilize the generative adversarial imitation learning (GAIL) model developed by \citet{ho2016generative}, a reinforcement learning variant of the generative adversarial network (GAN) developed by \citet{goodfellow2014generative}, used famously for video and image generation. We also use the large-scale pre-trained language model GPT-2 within the Generator of the GAIL architecture that allows for basic textual understanding and generation \citep{radford2019language}. Our novel implementation allows us to fine tune the language model using the expert empathetic trajectories for empathetic dialogue generation. For the optimization, we utilize proximal policy optimization (PPO) \citep{schulman2017proximal}.   

For a fair evaluation, we utilize an error metric commonly used in the field of natural language processing to evaluate langauge models. We evaluate our model over single-turn and multi-turn conversations using the perplexity and BLEU error metrics \citep{chen1998evaluation, papineni2002bleu}. Instead of testing our model's performance over common baselines for natural dialogue generation including COCO, ROCStories, and CommonGEN \citep{lin2014microsoft, lin2019commongen, mostafazadeh2017lsdsem}, we evaluate our model against others using our own empathetic dataset. The motivation for such a methodology is that the typical benchmarks used for conditional and unconditional text generation do not contain empathetic prompts or responses, and as a result, our model's effectiveness will not be tested. Moreover, our model does not accomplish the same tasks that would be measured in these text generation baseline datasets.

We make several contributions: (1) we propose an application of the generative adversarial imitation learning architecture for text generation, (2) we fine tune the large-scale pre-trained language model GPT-2 using expert empathetic actions to generate empathetic dialogues, (3) we show that our novel architecture outperforms similar text generation models in singe-turn and multi-turn empathetic dialogues using the perplexity and BLEU error metrics. 

\section{Related Work}
Artificial empathy is a large field, however, research into intelligent conversation agents with empathy is scarce. One of the most famous proprietary chatbot systems available to consumers is WoeBot developed by \citet{fitzpatrick2017delivering}. WoeBot is an intelligent conversation agent that allows patients to converse in a structured set of prompts to deal with various mental health issues. The intelligent agent is certified in cognitive behavioral therapy (CBT). Although it does not provide natural context-aware text generation, as the set of prompts are relatively static with respect to the subject of the conversation, WoeBot is able to discover and respond to human emotions. 

\citet{fung2018empathetic} is one of the first papers to introduce the application of deep reinforcement learning for empathetic dialogue generation. Their work describes a reinforcement learning variant of a traditional Seq2Seq model used for dialogue generation. Instead of defining a reward function for empathy, they seek to categorize each human prompt into an emotion class. They classify each prompt into an emotion class by the words and emoticons used in each utterance, and then respond using a set of predefined responses based on the discovered emotion class. They train their deep reinforcement learning architecture using Proximal Policy Optimization (PPO). 

\citet{wu2020textgail} is perhaps the most similar to our proposed research. They propose the application of a generative adversarial imitation learning model for traditional text generation. They utilize pre-trained language models including the RoBERTa architecture for text generation within the GAIL architecture \citep{liu2019roberta}. Their model shows improvement over previous work on text generation of image captions and baseline text generation tasks including COCO, CommonGEN, and ROCStories. 

\section{Model Architecture}

\begin{figure}
    \centering
    \includegraphics[width=\textwidth]{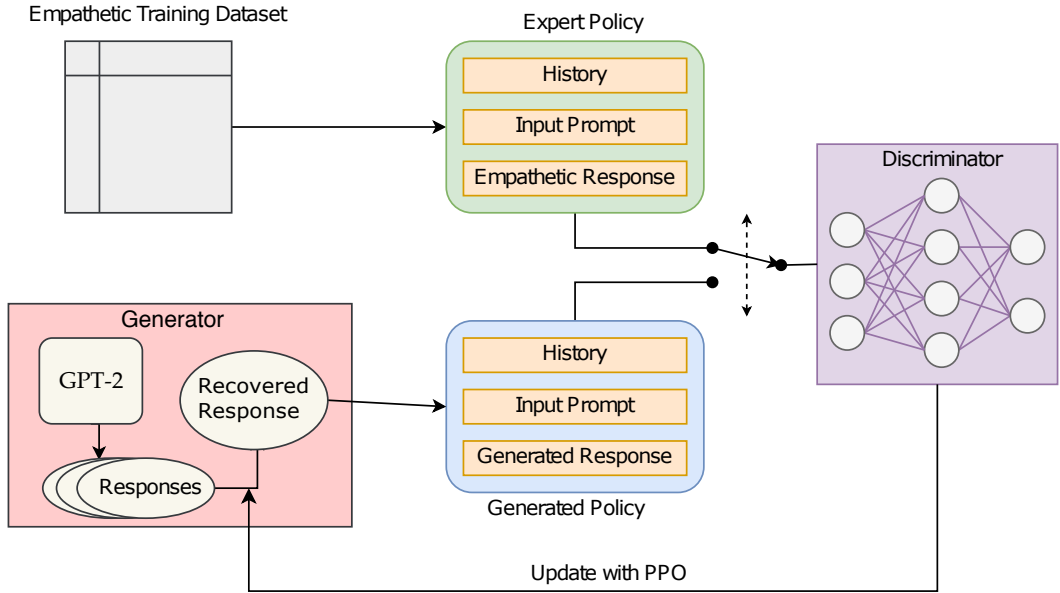}
    \caption{Model Architecture}
    \label{fig:arch}
\end{figure}

In this section, we describe the architecture of our proposed model. We describe our novel fine tuning approach on pre-trained language models using expert empathetic actions, as well as the structure of the Generator and Discriminator models within our generative adversarial imitation learning architecture. A visualization of the overall architecture of our model is described in Figure \ref{fig:arch}.

For dialogue generation, we modify the generative adversarial imitation learning architecture similar to \citet{wu2020textgail}. Generative adversarial imitation learning is a model-free imitation learning architecture able to directly imitate near-optimal expert trajectories in high dimensional, complex environments. The GAIL architecture consists of a generator G and discriminator D, where the generator seeks to generate responses similar to empathetic human responses, while the discriminator seeks to distinguish between the generated responses and the expert empathetic responses and propagate a reward signal back to the generator. Note that the reward signal here is not the reward function that a reinforcement learning agent must maximize, but it is a measure of similarity, defined as `occupancy measure' in the paper. As a result, the GAIL architecture does not ever need to discover the exact reward function that a typical reinforcement learning agent would maximize, which suits our goal of empathetic dialogue generation. 

The generator of the GAIL model performs imitation learning by constructing an environment where the inverse reinforcement learning problem is the dual of the reinforcement learning problem. That is, we define the reinforcement learning problem as

\begin{equation}
    RL(c) = \arg\min_{\pi \in \prod} - H(\pi) + \mathbb{E}[c(s,a)],
\end{equation}
where $\pi \in \prod$ is a recovered policy, $H(\pi)$ is the $\gamma-$distributed entropy of policy $\pi$, and $c(s,a)$ is the cost for state $s$ and action $a$.
Similarly, we then define the $\psi$-regularized maximum entropy inverse reinforcement learning problem as 
\begin{equation}
    IRL_\psi(\pi^E) = \arg\max_{c\in \mathbb{R}^{s\times a}} -\psi(c) + \bigg( \min_{\pi \in \prod} -H(\pi) + \mathbb{E}_\pi[c(s,a)] \bigg) - \mathbb{E}_{\pi^E}[c(s,a)],
\end{equation}
where $\psi$ is a regularization term and $\pi^E$ is the optimal policy. Note that $\psi$-regularized inverse reinforcement learning implicitly seeks a policy where its occupancy measure $\rho$ is close to the expert. 

The generated policy of the GAIL is then 
\begin{equation}
    RL \circ IRL_{\psi}(\pi^E) = \arg\min_{\pi \in \prod} -H(\pi) + \psi^*(\rho_\pi - \rho_{\pi^E}),
\end{equation}
where $\rho_\pi$ is the occupancy measure of the recovered policy and $\rho_{\pi^E}$ is the occupancy measure of the expert policy. We can select a regularizer that allows us to implement a more complex class of cost functions using neural networks. In our implementation, we define our cost regularizer as 

\begin{equation}
    \psi_{GA}(c) := \left\{ \begin{array}{ll}
            \mathbb{E}_{\pi^E}[g(c(s,a))] & \text{if } c < 0 \\
            +\infty &\text{otherwise}
        \end{array} \right. 
\end{equation}
where
\begin{equation}
        g(x) = \left\{ \begin{array}{ll}
            -x-\log(1-e^x) & \text{if } x < 0 \\
            +\infty &\text{otherwise.}
        \end{array} \right. 
\end{equation}

The motivation behind this specific cost regularizer is because the convex conjugate $\psi^*$ is the optimal negative log-loss of the binary classification problem of distinguishing between state-action pairs of $\pi$ and $\pi_E$:
\begin{equation}
    \psi^*_{GA}(\rho_\pi - \rho_{\pi_E}) = \max_{D \in (0,1)^{S \times A}} \mathbb{E}_\pi[\log D(s,a)] + \mathbb{E}_{\pi_E}[\log(1-D(s,a))] \label{eq:disc},
\end{equation}
which is also exactly the same as the cost function of the discriminator network of a traditional generative adversarial network (GAN). 

For dialogue generation, we replace the state $s$ within the GAIL with a two-part input and its corresponding action $a$ with the target response. This two-part input consists of (1) an optional history of the earlier prompts and responses in the conversation and (2) a required input prompt. We also optimize the policy generation within the generator of the GAIL with PPO instead of trust region policy optimization (TRPO) \citep{schulman2015trust}, which was used in the original paper. The TRPO algorithm is a constrained optimization problem that ensures that the policy is not moving too far away from the starting point by using the KL-divergence. TRPO is a commonly used policy optimizer, but more recent research has shown newer policy optimizers that have lower variance and do not solely require small steps to convergence \citep{engstrom2019implementation}. PPO is a simpler, more effective policy optimization algorithm compared to TRPO. PPO tries to compute an update at each time step that minimizes the cost function while ensuring the deviation from the previous policy is small. Instead of a KL-divergence constraint like TRPO, PPO uses a KL-divergence penalty: \begin{equation}
    \max_\theta \sum_{n=1}^N \frac{\pi_\theta (a_t | s_t)}{\pi_{\theta_\text{old}} (a_t | s_t)} \hat{A}_t - C \cdot KL_{\pi_{\theta_{\text{old}}}}(\pi_\theta). 
\end{equation}

For the Generator network, we utilize the pre-trained language model GPT-2, a large-scale multipurpose language model with a transformer architecture developed by OpenAI with 117 million parameters trained on 40 GB of internet data. The nature of our modified GAIL model allows for the expert empathetic actions to fine tune the baseline language model for empathetic dialogue generation in a similar manner to the fine tuning process found in transfer learning. Essentially, instead of retraining the language model for empathetic dialogues, which would be computationally infeasible due to the size of the language model, the GAIL model treats all possible GPT-2 responses for a prompt as the set of possible actions to take in the state-space, and the architecture selects the response most similar with respect to the occupancy score of the optimal expert empathetic response. 

We describe the training process of our overall modified generative adversarial imitation learning algorithm in Algorithm \ref{alg:gail}. We also describe the training process of the generator network within the modified GAIL architecture in Algorithm \ref{alg:gen}. 

\begin{algorithm}
\caption{Modified GAIL Algorithm}
\textbf{Input:} Expert empathetic trajectories $\tau_E$ each with history $h_E$, prompt $p_E$ and response $r_E$; initial policy and discriminator parameters $\theta_0$, $w_0$
\begin{algorithmic}
\label{alg:gail}
\FOR{ $i = 0,1,2, \dots$}
\STATE Sample trajectories $\tau_i$ for discriminator $D$
\STATE Collect occupancy score $\rho_i$ from discriminator $D$ for all $\tau_i$
\STATE Update discriminator parameters from $w_i$ to $w_{i+1}$ using Eq. \ref{eq:disc}
\STATE Update generator parameters by taking policy step from $\theta_i$ to $\theta_{i+1}$ using PPO
\ENDFOR
\end{algorithmic}
\end{algorithm}

\begin{algorithm}
\caption{Generator Algorithm}
\textbf{Input:} Initial policy parameters $\theta_0$, Noise prior $p(z)$
\begin{algorithmic}
\label{alg:gen}
\FOR{ $i = 0,1,2, \dots$}
\STATE Sample noise sample $z_i \sim p(z)$
\STATE Generate possible responses using GPT-2 language model
\STATE Recover generated response using $z_i$ to form $\gamma_i$
\STATE Update parameters by taking policy step from $\theta_i$ to $\theta_{i+1}$ using PPO
\ENDFOR
\end{algorithmic}
\textbf{Output:}  Generated policy $\gamma_G$ each with history $h_G$, prompt $p_G$, and response $r_G$ 
\end{algorithm}

\section{Experiments} \label{exp}
In this section, we describe the datasets and implementation details of our experiments. 

\subsection{Datasets}\label{data}

\begin{figure}[ht]
    \centering
    \includegraphics[width=\textwidth]{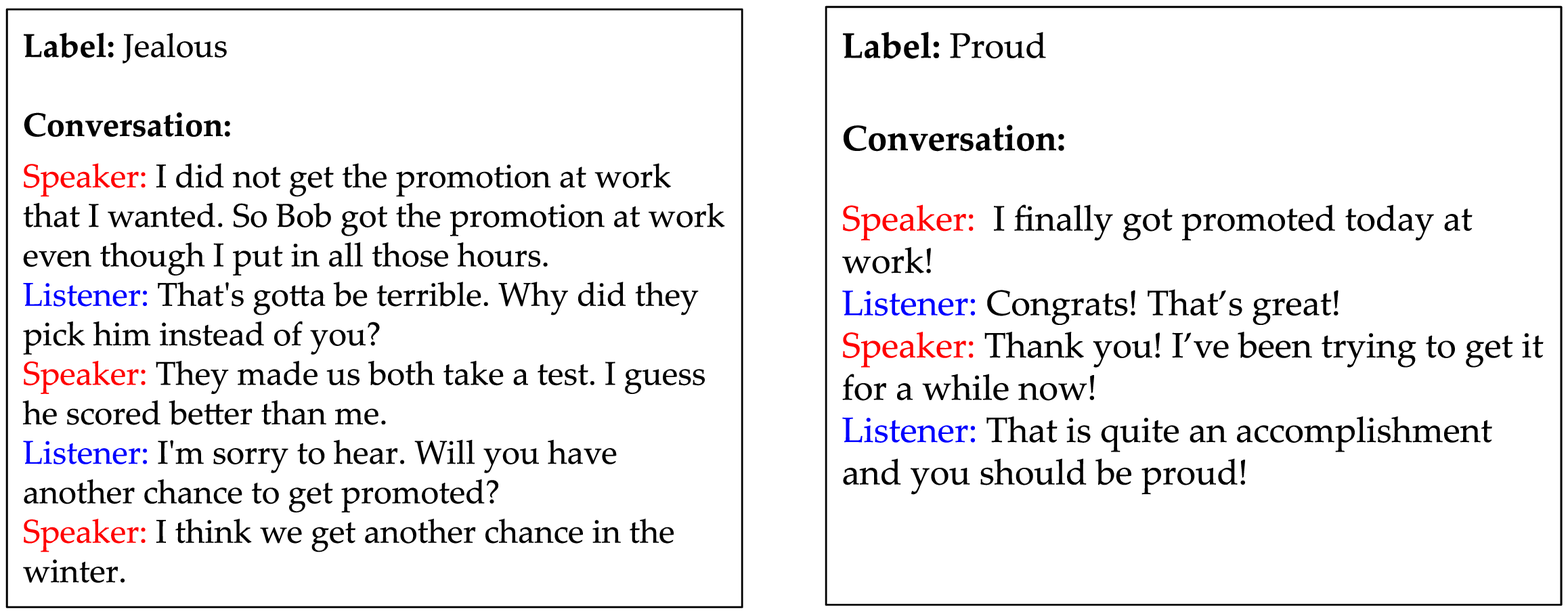}
    \caption{EmpatheticDialogues Example Data}
    \label{fig:emp}
\end{figure}

The datasets we used for the following experiments were selected to be clear examples of empathetic conversations, actions, or dialogues. Our dataset plays a larger role in the performance of our model compared to similar research because the empathetic nature of our model comes from imitating our dataset of expert empathetic trajectories. As a result, we chose our dataset very carefully in order to exemplify human empathetic actions in a broad setting. Note that all data we use in this paper is open source, anonymized, and does not contain any personally identifiable or offensive content.  

The first dataset we used is the EmpatheticDialogues dataset collected by Facebook AI Research \citep{rashkin2018towards}. The EmpatheticDialogues dataset consists of 25 thousand single-turn and multi-turn empathetic conversations. Each data sample contains (1) a label which describes the overall emotion of the conversation and (2) the transcript of a conversation between exactly two human speakers. These conversations were human-labeled and human-generated through crowd sourcing via Amazon Mechanical Turk. An example of the data is shown in Figure \ref{fig:emp}. 

\begin{figure}[ht]
    \centering
    \includegraphics[width=0.5\textwidth]{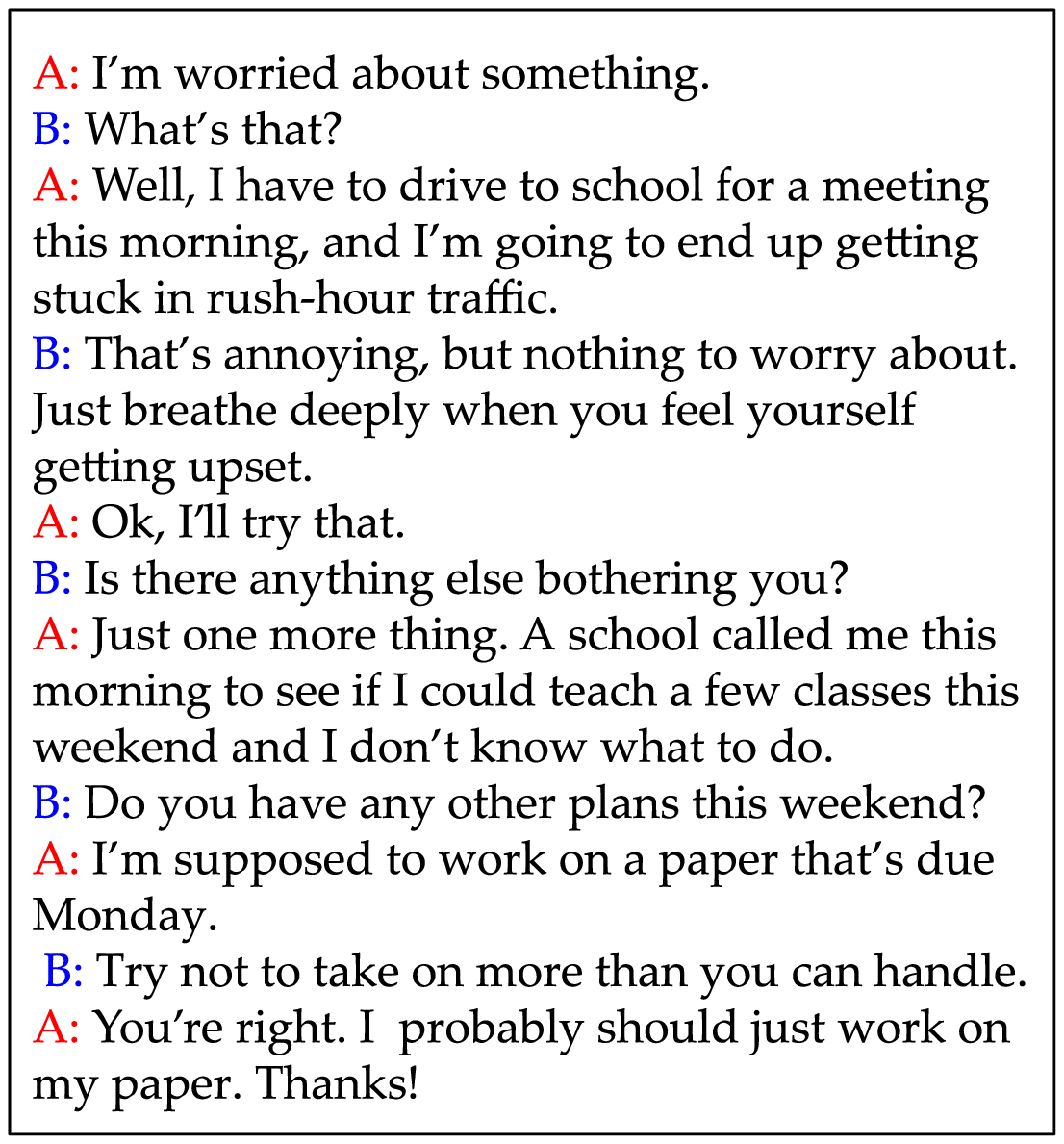}
    \caption{DailyDialog Example Conversation}
    \label{fig:daily}
\end{figure}

The second dataset we used is the DailyDialog dataset collected by \citet{li2017dailydialog} consisting of 13 thousand multi-turn empathetic conversations. The data labeling task was crowd sourced and human-labeled. We use this dataset as it provides longer example conversations sustained over a single topic. This allows our model to produce more context-aware and history-based dialogue generation, as it imitates the sustained flow of conversation evident in the DailyDialog dataset. A sample DailyDialog conversation is shown in Figure \ref{fig:daily}.

\begin{figure}[ht]
    \centering
    \includegraphics[width=\textwidth]{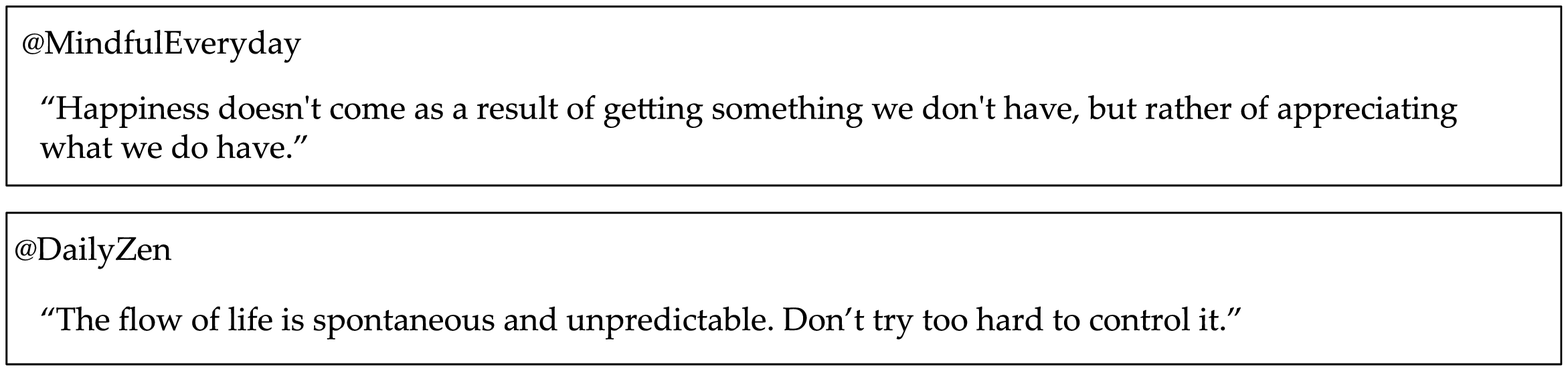}
    \caption{Example Empathetic Tweets}
    \label{fig:tweet}
\end{figure}

Finally, we also collect empathetic tweets as a datasource. We find that including empathetic tweets allow our model to offer advice or provide inspirational responses at certain points in the conversation, resulting in a more thoughtful and altruistic conversational agent. We collect the entire twitter histories of established uplifting Twitter accounts including @DalaiLama, @DailyZen, and @MindfulEveryday. Some examples of these tweets are shown in Figure \ref{fig:tweet}.

\subsection{Implementation}\label{imp}

\begin{table}
    \centering
    \small
    \begin{tabular}{l c c}
    \hline
        \bf{Hyperparameter} \hspace{5cm} & \bf{Value}  \\
    \hline
        Sample Batch Size \hspace{5cm} & 16 \\
        Generator Warmup Scheduler \hspace{5cm} & WarmupLinear \\
        Generator Warmup Steps \hspace{5cm} & 1000 \\
        Human Demo Ratio \hspace{5cm} & 0.3 \\
        Human Demo Ratio Warmup Steps \hspace{5cm} & 100 \\
        PPO Buffer Size \hspace{5cm} & 128 \\
        PPO Minibatch Size \hspace{5cm} & 8 \\
        PPO Epsilon \hspace{5cm} & 0.2 \\
        Discriminator Pre-Train Steps \hspace{5cm} & 200 \\
        Discriminator Optimizer \hspace{5cm} & AdamW \\
        Discriminator Optimizer Learning Rate \hspace{5cm} & $1 \times 10^{-4}$ \\ 
        Weight Decay \hspace{5cm} & 0.01 \\
        Layer Normalization Epsilon \hspace{5cm} & $1 \times 10^{-5}$ \\
        Activation Function \hspace{5cm} & tanh \\
        Epochs \hspace{5cm} & 750 \\
        Batch Size \hspace{5cm} & 32 \\
        Validation Frequency \hspace{5cm} & 10\\
    \hline
    \end{tabular}
    \caption{Model Hyperparameter Values}
    \label{tab:hyperparameter}
\end{table}

Our implementation of our modified generative adversarial imitation learning architecture uses a neural network architecture with two hidden layers of 100 units each, with $\tanh$ nonlinearities in between for the generator and discriminator. The pre-trained model weights for the Generator network is the base GPT-2 model. A majority of our implementation was completed using the PyTorch Python framework for Python 3.7.7 \citep{pytorch}. All external frameworks used in our implementation including GPT-2 and PyTorch are accessed and provided as open-source assets. 

To create the expert empathetic trajectories from the dataset, we must pre-process the conversation data. For the EmpatheticDialogues and DailyDialog datasets, we create an expert trajectory for each turn of the conversation, where a turn is defined as a prompt speaker from one speaker and a response utterance from another speaker. If the data sample is a single turn conversation, then we construct an expert empathetic trajectory sample where the history $h_i$ is an empty string, the input prompt $p_i$ is the first speaker's utterance, and the optimal response $r_i$ is the second speaker's utterance. For multi turn conversations, we stagger the generation of expert trajectories such that we create a trajectory for the first turn in the conversation with no history $h_i$, but the first two utterances as prompt $p_i$ and response $r_i$. For every turn after the first, we concatenate all previous turns before the current turn and store the utterances as the history $h_i$, the current turn's first speaker utterance as the prompt $p_i$ and the current turn's second speaker utterance as the response $r_i$. 

For each conversation in either the EmpatheticDialogues or DailyDialog dataset, there is at least one generated empathetic trajectory. In general, there are usually more generated for each data sample, since most conversations in these datasets are multi turn: EmpatheticDialogues has an average of 2.3 turns per conversation; DailyDialog has an average of 4.8 turns per conversation. 

For the empathetic tweets that are not in conversation format, we store no history $h_i$ but store the tweet text as both the prompt $p_i$ and the response $r_i$. For each tweet, there is exactly one generated expert empathetic trajectory. In total, we generate 208600 expert empathetic trajectories: 79353 from EmpatheticDialogues, 61935 from DailyDialog, and 67312 from empathetic tweets.

We train our model over 750 epochs with a batch size of 32, a validation frequency of every 10 epochs, and a human demonstration mix ratio set at 0.3 at the start of training which was configured to decrease linearly throughout training. Our train-test-validation split was 70-20-10 respectively. Our generator network optimizes our policy through a custom minimal implementation of PPO with a minibatch size of 8 and an epsilon value of 0.2. We evaluate our error using the perplexity and BLEU error metrics commonly used to measure the effectiveness of language models in the field of natural language processing. 

We describe additional hyperparameter selections for our model implementation in Table \ref{tab:hyperparameter}. All hyperparameter choices were either chosen empirically or left as default as a result of running various hyperparameter tuning iterations. We train and evaluate our model using the Google Colaboroatory GPU-enabled runtime with a 2496 CUDA core Tesla K80 GPU with 12GB GDDR5 VRAM and a single core multi thread 2.3 gigahertz Intel Xeon CPU. The training process of 750 epochs completed in roughly four compute hours. 

\section{Results} \label{res}

\begin{figure}
\centering
\begin{minipage}{.5\textwidth}
  \centering
  \includegraphics[width=\linewidth]{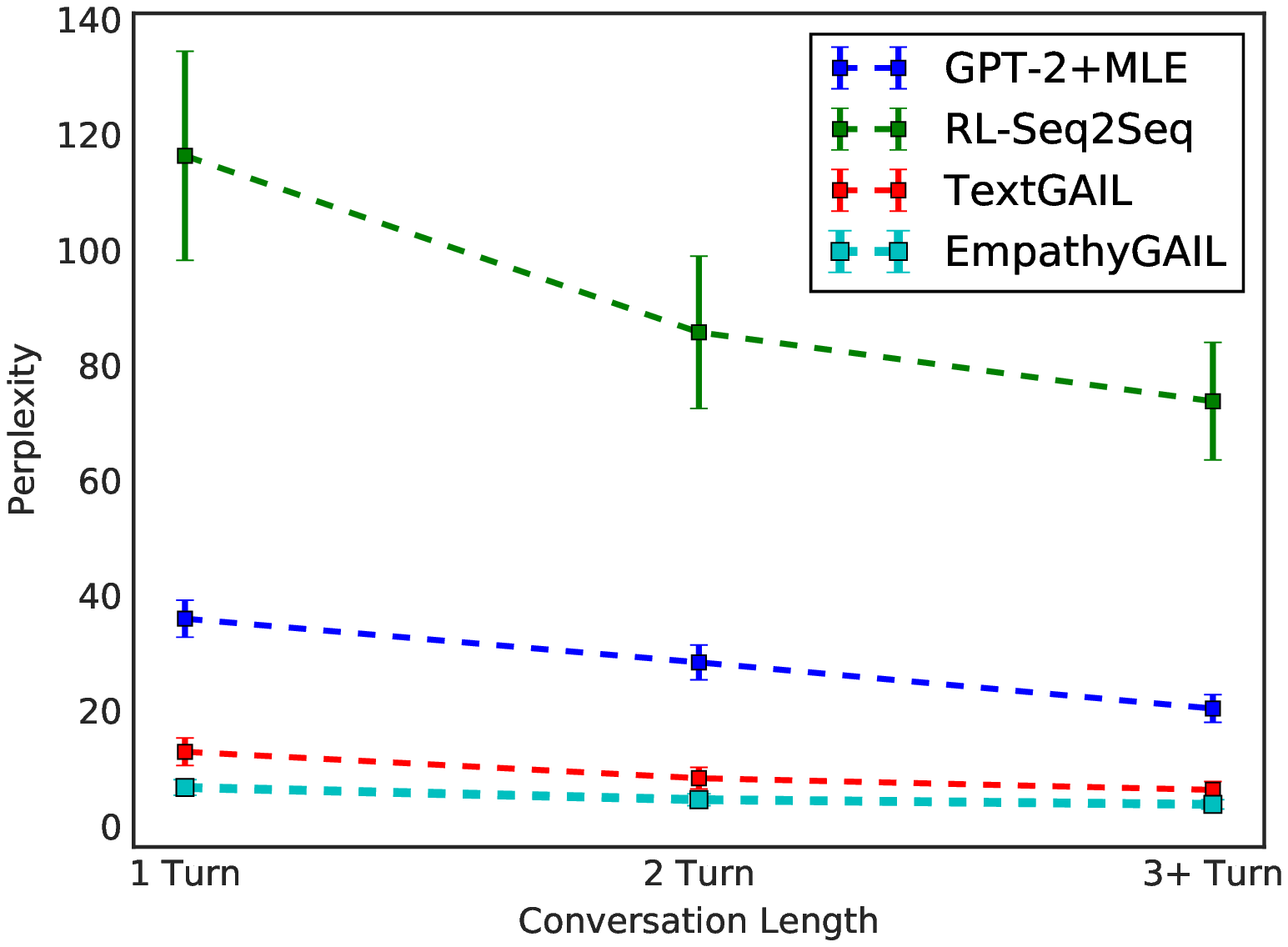}
  \captionof{figure}{Perplexity scores}
  \label{fig:perp}
\end{minipage}%
\begin{minipage}{.5\textwidth}
  \centering
  \includegraphics[width=\linewidth]{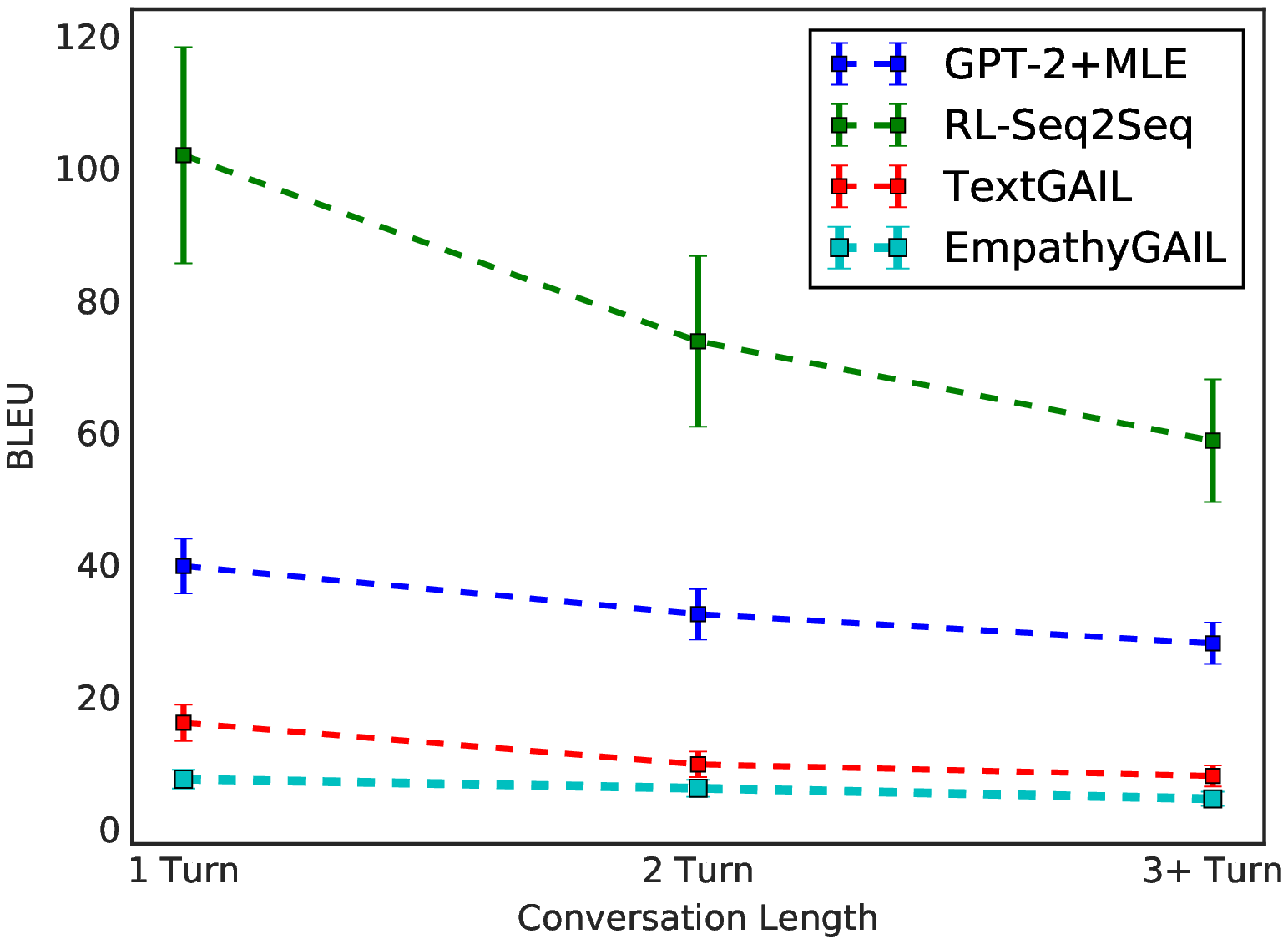}
  \captionof{figure}{BLEU scores}
  \label{fig:bleu}
\end{minipage}
\end{figure}

To evaluate the effectiveness of our model, we collect error metric results for our model against baseline dialogue generation models. 
The error metrics we use throughout our experiments are the perplexity and BLEU error metrics. In a general sense, perplexity measures the human-readability of a generated text, where a low perplexity score corresponds to an easily understandable output. The perplexity score $P$ for a language model $p_M($next word $w|$ history $h)$ on a test set $T = \{ w_1, \dots, w_t\}$ is $$P(p_M) = \frac{1}{(\prod_{i=1}^T p_M(w_i|w_1\dots w_{i-1}))^{\frac{1}{t}})}.$$ The Bilingual Evaluation Understudy Score (BLEU) error metric measures the similarity of a candidate sentence to a target sentence. Here, we will measure the BLEU scores on generated responses to optimal empathetic responses in the test set. 

\begin{table}[ht]
    \centering
    \begin{tabular}{ccccc} \toprule
    {Metric} & {Model} & {1 Turn} & {2 Turn} & {3+ Turn} \\[0.3cm] \midrule
     \multirow{4}{*}{Perplexity} & GPT-2 + MLE  & 36.61 $\pm$ 3.21 & 29.03 $\pm$ 3.02 & 21.02 $\pm$ 2.41 \\
    & RL-Seq2Seq  & 116.96 $\pm$ 18.14 & 86.31 $\pm$ 13.23 & 74.35 $\pm$ 10.20 \\
    & TextGAIL  & 13.51 $\pm$ 2.39 & 8.93 $\pm$ 1.88 & 6.90 $\pm$ 1.45\\
    & $\textbf{EmpathyGAIL}$  & $\bf{7.32 \pm 1.35}$ & $\bf{5.20 \pm 1.03}$ & $\bf{4.38 \pm 0.81}$ \\\midrule
    \multirow{4}{*}{BLEU} & GPT-2 + MLE  & 40.34 $\pm$ 4.15 & 33.05 $\pm$ 3.82 & 28.64 $\pm$ 3.12 \\
    & RL-Seq2Seq  & 102.50 $\pm$ 16.35 & 74.34 $\pm$ 12.91 & 59.30 $\pm$ 9.29 \\
    & TextGAIL  & 16.63 $\pm$ 2.75 & 10.34 $\pm$ 1.94 & 8.58 $\pm$ 1.59 \\
    & $\textbf{EmpathyGAIL}$  & $\bf{8.10 \pm 1.42}$ & $\bf{6.72 \pm 1.29}$ & $\bf{5.12 \pm 1.07}$ \\\bottomrule
\end{tabular}
    \vspace{0.2cm}
    \caption{Perplexity and BLEU Error Results for 1 Turn, 2 Turn, and 3+ Turn Conversations}
    \label{tab:results}
\end{table}

We test our model against three baselines. For each baseline, due to the nature of our problem, we collect perplexity error by training and evaluating these models on our empathetic dataset as opposed to a common benchmark dataset. For all baselines, we utilize the default hyperparameters described in their respective implementations. All baselines we select are dialogue generation models. Our first baseline is a base GPT-2 dialogue generation instance fine tuned to our dataset through maximum likelihood estimation (MLE). Our second baseline is the reinforcement learning variant of the Seq2Seq architecture developed by \citet{fung2018empathetic} where they classify input prompts into emotion classes to respond empathetically. Our final baseline is the TextGAIL model developed by \citet{wu2020textgail} that utilizes the RoBERTa pre-trained language model within the architecture of a GAIL model for dialogue generation. 

We test our model against the baselines using the BLEU and perplexity error metrics for a single turn, 2 turn, and 3+ turn conversations. That is, a single turn conversation contains no history; a 2 turn conversation contains the history of a single turn; and a 3+ turn conversation contains the history of multiple turns. We provide our results in Table \ref{tab:results} along with error bars with respect to various iterations using different random seeds. We provide a visualization of perplexity and BLEU error over conversation length for each model in Figure \ref{fig:perp} and \ref{fig:bleu} respectively. Our BLEU and perplexity error metric results show that our model, $\emph{EmpathyGAIL}$, provides significant improvement over baselines. 

In order to display the empathetic abilities of our approach, we also provide some example responses from our model. Each example response corresponds to an input prompt and an optional history. Table \ref{tab:out} describes a few of these results. 

\begin{table}[ht]
    \centering
\begin{tabular}{ll} \toprule
     & {Example Results} \\ \midrule
    $\textbf{History}$ & I moved to LA a few years ago and I never had the chance to visit home. \\
    $\textbf{Prompt}$ & I'm going to see my parents soon! \\
    $\textbf{Generated Response}$ & That’s awesome! I hope you have a good time. \\\midrule
    $\textbf{History}$ & None \\
    $\textbf{Prompt}$ & He was a very good friend of mine. Last month, he passed away. \\
    $\textbf{Generated Response}$ & I am very sad to hear that he is no longer alive.\\\midrule
    $\textbf{History}$ & I was in a drive-thru line and someone cut in front of me. \\
    $\textbf{Prompt}$ & I was furious and blew up at him. \\
    \multirow{2}{*}{$\textbf{Generated Response}$} & Sometimes it makes more sense to just walk away. \\
    & You cannot control others, you can only control yourself. \\\midrule
    $\textbf{History}$ & None \\
    $\textbf{Prompt}$ & I had a job interview today and I think it went really well. \\
    $\textbf{Generated Response}$ & Wonderful!\\\bottomrule
\end{tabular}
    \vspace{0.1cm}
    \caption{Example Model Results}
    \label{tab:out}
\end{table}

\section{Conclusion} \label{conclusion}

We propose a generative adversarial imitation learning architecture for empathetic dialogue generation. Our key contributions include a modified GAIL architecture for dialogue generation, a fine tuning approach using expert empathetic trajectories on large-scale pre-trained language models, and our results show improved perplexity and BLEU scores over single turn and multi turn conversations. Our work shows that deep imitation learning models can accurately and effectively provide context-aware, empathetic, and natural dialogue generation.

Our results show an average perplexity error decrease of 80\% and a BLEU error decrease of 81\% compared to a base GPT-2 model fine tuned to our dataset through MLE. Our results also show an average perplexity error decrease of 42\% and a BLEU error decrease of 44\% compared to a generative adversarial imitation learning model for text generation \citep{wu2020textgail}.

There are a few limitations of our model. We find that at certain times, our model will begin to generate fake experiences or stories in an attempt to connect with the input prompt. This is because the input empathetic datasets we use are entirely human generated such that a human responder can empathetically respond to a human prompt by narrating their own prior experiences. Since our architecture seeks to imitate these empathetic responses, we find that our model does the same. Also the GPT-2 pre-trained language model we use within the generator of the GAIL is trained on datasets which contain biases or factual inaccuracies, and as a result will be reflected in the output of our model. 

\section{Future Work and Impacts} \label{impacts}

We hope to use additional sources of empathetic data which are human-labeled but may not be human-generated to ensure that our model does not generate fake experiences or stories to connect with the input prompt. We also hope to implement more powerful pre-trained language models like the recently developed GPT-3 model \citep{brown2020language}. Another direction we would like to take our research is developing a more personalized sense of empathy for each user. Empathy is individual-specific, so we hope to implement a set of calibrating questions that gauge the perception of empathy for a user prior to providing empathetic dialogue. 

We forsee our impacts as a tool for patients suffering with mental health, anxiety, or depression to converse with an empathetic chatbot at any time. Before making our developments available for public use, we will have to complete rigorous training and testing to ensure that the chatbot provides help and not harm. There may be serious negative societal impacts if adequate testing is not completed, so we will heavily focus our future research into ensuring the safety and effectiveness of our work.  

\label{refs}
\bibliographystyle{apalike}
\bibliography{ref}


\end{document}